\documentclass[twoside,twocolumn]{article}

\usepackage{blindtext} 

\usepackage[sc]{mathpazo} 
\usepackage[T1]{fontenc} 
\usepackage{microtype} 

\usepackage[english]{babel} 

\usepackage[hmarginratio=1:1,top=32mm,columnsep=20pt]{geometry} 
\usepackage[hang, small,labelfont=bf,up,textfont=it,up]{caption} 
\usepackage{booktabs} 

\usepackage{lettrine} 
\usepackage{subcaption}

\usepackage{enumitem} 
\setlist[itemize]{noitemsep} 

\usepackage{abstract} 

\usepackage{titlesec} 
\renewcommand\thesection{\Roman{section}} 
\renewcommand\thesubsection{\roman{subsection}} 
\titleformat{\section}[block]{\large\scshape\centering}{\thesection.}{1em}{} 
\titleformat{\subsection}[block]{\large}{\thesubsection.}{1em}{} 

\usepackage{titling} 

\usepackage[hyphens]{url}  
\usepackage[pdftex,bookmarksnumbered,hidelinks,breaklinks]{hyperref}
\makeatletter
\g@addto@macro\UrlBreaks{\do\-}
\makeatother

\usepackage{graphicx}

\providecommand{\keywords}[1]
{
  \small	
  \textbf{\textit{Keywords---}} #1
}

\title{Practical data monitoring in the internet-services domain} 
\author{%
\textsc{Nikhil Galagali}\\[1ex]
\normalsize \href{mailto:nikhilg18@gmail.com}{nikhilg18@gmail.com} 
}
\date{\today} 
\renewcommand{\maketitlehookd}{%
\begin{abstract}
Large-scale monitoring, anomaly detection, and root cause analysis of metrics are essential requirements of the internet-services industry. To address the need to continuously monitor millions of metrics, many anomaly detection approaches are being used on a daily basis by large internet-based companies. However, in spite of the significant progress made to accurately and efficiently detect anomalies in metrics, the sheer scale of the number of metrics has meant there are still a large number of false alarms that need to be investigated. This paper presents a framework for reliable large-scale anomaly detection. It is significantly more accurate than existing approaches and allows for easy interpretation of models, thus enabling practical data monitoring in the internet-services domain.
\\
\\
\keywords{Time series analysis; anomaly detection; machine learning; state space modeling}
\end{abstract}
}

\begin{document}

\maketitle

\section{Introduction}
With the emergence of a large number of internet-based businesses, ranging from e-commerce sites to social media and ride-hailing platforms, more and more companies are serving their customers through the internet. Backing these web services are cloud-based software systems that generate petabytes of data each day. Data in this context are typically metrics reflecting the fidelity of the applications being served, infrastructure health metrics, and business metrics summarizing revenue and user engagement of the service. Being able to continuously monitor different metrics generated by the service is a key requirement of today's internet-services industry. Periodic tracking of metrics for any abnormal behavior helps companies ensures reliable upkeep of their services and rapid mitigation of any incident that could detrimentally affect user experience, and ultimately lead to revenue losses.

\begin{figure*}[h]
\includegraphics[width=\textwidth]{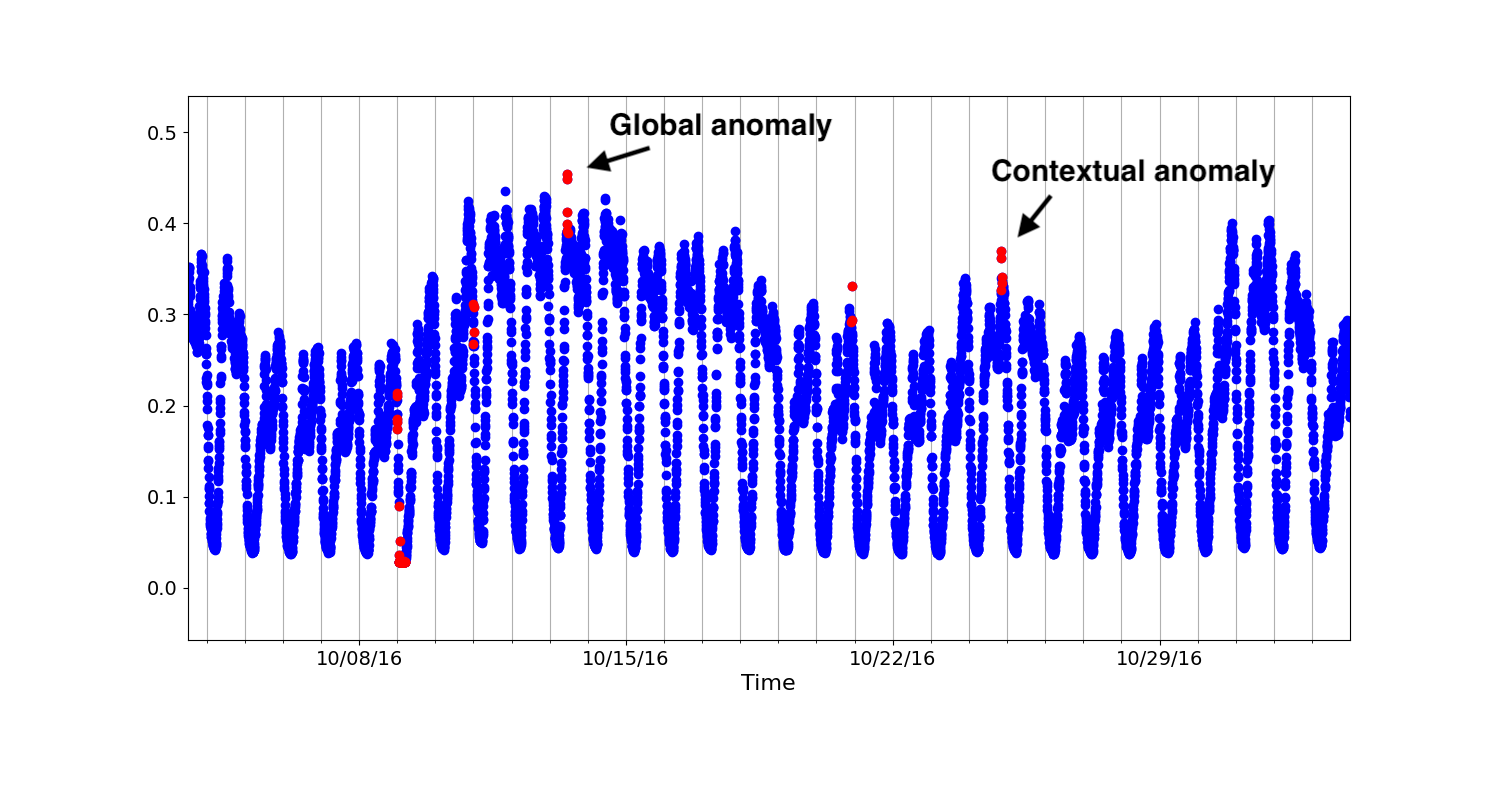}
\caption{Time series with global and contextual anomalies}
\label{fig:5}  
\end{figure*}

Anomaly in a dataset refers to data points that deviate from normal behavior. Anomalies can occur due to a number of reasons, such as malicious actors, system failure, or change in user behavior. Further, as described in the review article \cite{chandola2009}, anomalies are typically contextual. A data point that would be considered anomalous for a given context might not be an anomaly in another case (Figure \ref{fig:5}). Anomaly detection is the process of detecting any anomalous behavior and communicating all relevant information to the concerned human operators. In the internet services domain, typically metrics are tracked longitudinally in time. As a result, the anomaly detection that is most often of interest is over time-series data. Historically, time series anomaly detection in the internet companies has been done using rule-based approaches. These rule-based approaches have generally involved a human expert specifying a threshold beyond which an alert is triggered. The thresholds are set in terms of the raw values of the metrics or simple statistical summaries like mean or standard deviation of the metrics. Although simple to develop and follow, these rule-based anomaly detection approaches are fraught with high false positive rates. Further, even in cases where reliable rules have been developed, the sheer size of the number of metrics being tracked meant that these approaches haven't been scalable. As a result, in recent years, many companies have adopted machine learning based approaches for large-scale automatic time series anomaly detection \cite{yahoo2015,twitter2017,microsoft2019,alibaba2018,luminol,google2017}.

Machine learning based anomaly detection is indispensable for large-scale monitoring of metrics. Broadly, ML-based anomaly detection approaches attempt to build an understanding of the metric time series and try to use the developed model to classify future points as anomaly or not. ML-based anomaly detection in internet services comes with its own set of challenges. Principal among them is absence of normal/anomaly labels on historical data. Given that each organization can be monitoring thousands of metrics at any given point of time, it is humanly impossible to have ground truth labels for data points. Even in cases where labels are available, due to the continuously changing characteristics of metric time series and the nature of anomalies, using past labels for training machine learning models is typically unreliable. As a result, a key need of the anomaly detection approaches in this domain is that they be unsupervised, i.e., not be trained on past anomaly labels. Further, the growing complexity of web applications has meant that metrics that are monitored could have different characteristics. This could mean some are seasonal whereas others are not, some are continuous valued while others take mixed continuous and discrete values etc. Therefore, it is critically important that any approach adopted is generalizable and works across a range of use cases and metric types. Finally, a third major requirement of any approach used for anomaly detection in the internet-services domain is that it be easily interpretable. With millions of metrics being tracked continuously, in spite of high-quality models being used today for time series anomaly detection, there usually are a few alerts generated periodically, majority of which are false alarms. These alerts are then passed to domain experts who need to investigate the alerts for root causes. Having a model that is easily interpretable is important to be able to sift through all anomalies and focus on the most critical ones. A recent trend in time series anomaly detection has been to develop deep learning models. These deep learning based models have been shown to outperform classical time series models in some cases. However, their use has also meant that the model results have become hard to interpret. Practically, this has meant that anomaly detection users develop  alert fatigue and in some cases fail to recognize critical concerns in a timely fashion. 

This paper presents a structural time series framework for time series anomaly detection. Structural time series models are state space models, which enable modeling time series as a combination of different components. Each component encapsulates a different feature of the time series data. We show that by combining the structural time series modeling with rigorous model selection over a range of carefully chosen suite of models, one can obtain a highly accurate and interpretable anomaly detection technique for the internet-services domain. Our approach is generalizable to myriad different time series types and is unsupervised and adaptive to changing time series characteristics. 

The paper consists of five sections. In the next section, we outline related previous work on time series anomaly detection in the internet-services domain. Section 3 describes our methodology to reliable anomaly detection. Section 4 presents the experimental evaluation and comparison of our approach to other popular times series anomaly detection approaches in the domain. Finally, we conclude and provide directions for future work.

\section{Related Work}
Anomaly detection is a topic of high interest in the internet services industry and a number of methods have been developed to tackle the problem. The approaches adopted have ranged from traditional statistical techniques such as ARIMA modeling and SVD to more modern deep learning based solutions. Broadly the available solutions can be divided into supervised and unsupervised anomaly detection approaches.

Opprentice is a supervised learning framework that makes use of random forest classifier to detect anomalies \cite{opprentice2015}. EGADS is a generic framework for anomaly detection released by Yahoo Inc that made use of a number of traditional statistical approaches to model the time series and then subsequently used a classifier to identify the most interesting anomalies \cite{yahoo2015}. Google Inc. developed a deep learning based approach to detect anomalies and showed promising results on their own dataset. In general, supervised learning based approaches are limiting because they rely on the availability of a continuous stream of anomaly labels which is difficult to collect in practice.

At the same time, a number of unsupervised anomaly detection approaches have also been developed in recent years. Twitter-AD is a commonly used anomaly detection tool used in the industry that makes use of STL for smoothing the time series and ESD for subsequent outlier detection \cite{twitter2017}. LinkedIn developed Luminol \cite{luminol} that segments the time series into chunks and determines anomaly scores for the segments based on their frequency of occurrence. In 2018, \cite{alibaba2018} developed DONUT an unsupervised anomaly detection approach using variational auto encoders for seasonal KPIs. Inspired by visual saliency detection, Microsoft proposed a spectral residual technique for anomaly detection \cite{microsoft2019}. More recently \cite{tsbert2021} developed a technique for anomaly detection that leverages BERT, a neural network model popular for natural language processing, for time series modeling and produces very high accuracy on the industry standard AIOPS dataset. Even though the paper describes their approach as unsupervised, it relies on knowing anomaly labels for cleaning the training data used for model building. Although these recent deep learning based unsupervised models have led to steady improvement in detection accuracy, they are hard to interpret and thus are often challenging to rely on in practical settings.

\section{Methodology}
The goal of time series anomaly detection is, given a times series $X_{i=1..T}$ of univariate data points, predicting whether the data points are anomaly or normal. In the present use case, data points are obtained in a streaming fashion. The granularity of data points can typically range from minutely, hourly to daily observations. The objective therefore when a new observation $X_{t}$ comes in is to predict reliably if it should be tagged as an anomaly or not. If the data point is deemed to be an anomaly, an alert in the form of an email or a page is sent to the concerned recipients. 

\begin{figure}[h]
\centering
\includegraphics[width=\linewidth]{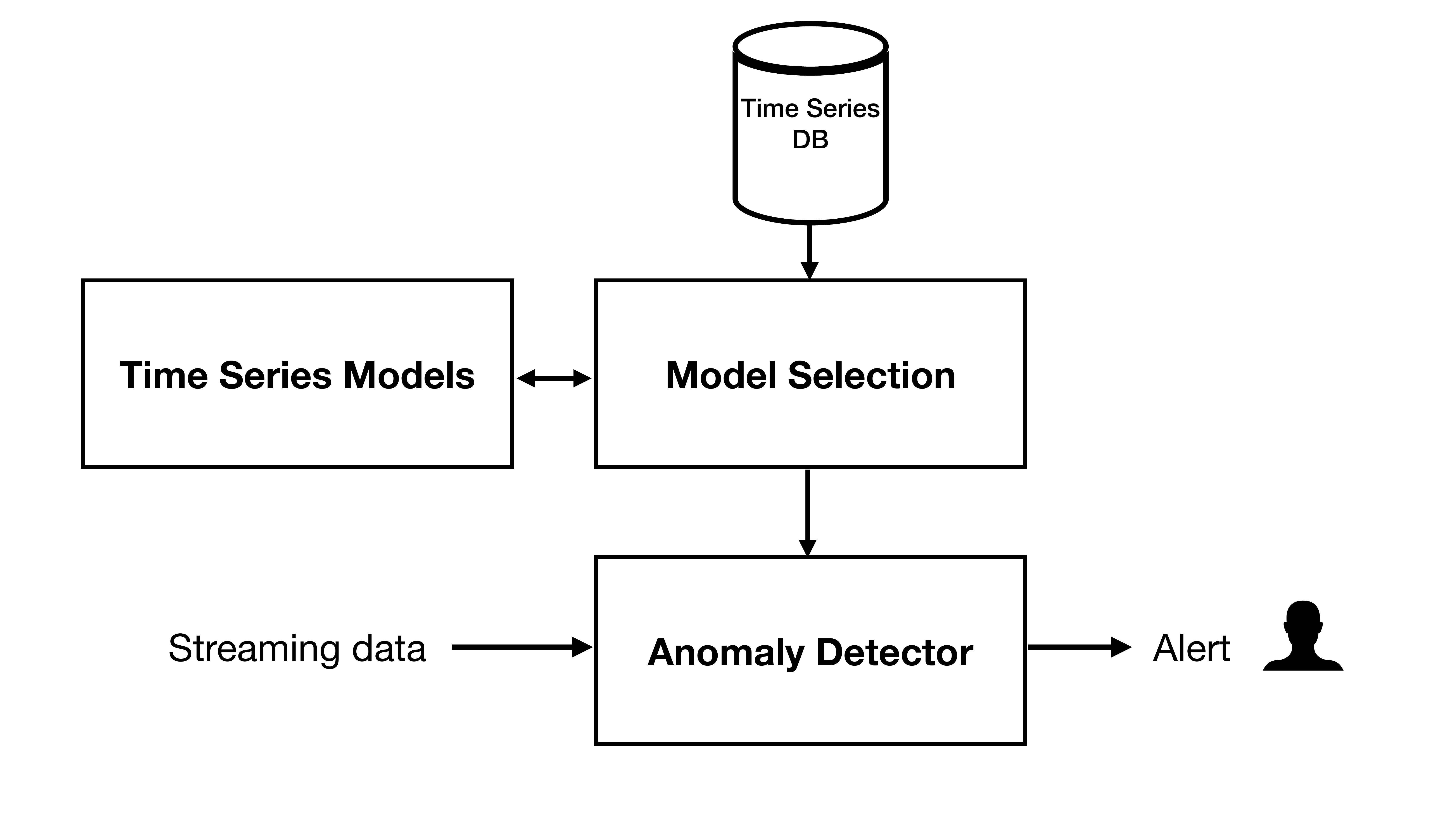}
\caption{System architecture of anomaly detection}
\label{system_arch}
\end{figure}

Our anomaly detection framework consists of three components: a time series modeling component, a model selection engine, and an anomaly detection module. As shown in Figure~\ref{system_arch}, the user of the framework configures a metric for anomaly detection. This involves defining the source of the time series metric. Once a metric is configured, a model selection engine evaluates a suite of models using  historical data on the metric in their ability to predict future values. We use an 80:20 split of available data for training and validation, and use mean square prediction error over the validation dataset to select the model with the lowest prediction error. Once the model is selected, it is used for all subsequent prediction and anomaly detection of the metric. As new data points of the metric stream into the system, the corresponding model and historical data of the metric are used to evaluate if the data point is an anomaly.

\subsection{Time series modeling}
A core requirement of our anomaly detection framework is the time series modeling capability. As described earlier, we adopt an unsupervised approach to anomaly detection. Consequently, we do not train the time series models to directly predict anomaly labels. Rather, the time series models are trained on historical data to predict future data points of a metric. We adopt the state space modeling approach to learn metric-specific models \cite{koopman2012}. State space modeling of time series is a general framework for modeling time series that consists of a hidden state $\alpha_{t}$ and observed data point $y_{t}$ at each time point $t$ related through the following equations: 

\begin{eqnarray}
y_{t}=Z_{t}\alpha_{t}+d_{t}+\epsilon_{t} \\
\alpha_{t}=S_{t}\alpha_{t-1}+c_{t}+R_{t}\eta_{t}
\end{eqnarray}

Here, $y_t$ is the observation vector, which in our case is univariate, $\alpha_{t}$ is the state vector at time $t$, $Z_t$, $S_{t}$, and $R_{t}$ are time varying matrices, $d_{t}$ and $c_{t}$ are inputs, and $\epsilon_{t}$ and $\eta_{t}$ are noise parameters. The central advantage of using the state modeling framework is that most commonly used classical time series models like regression models, Holt-Winters method, ARMA, ARIMA, seasonal ARIMA etc. are instances of the state space framework. In fact, even neural network models like recurrent neural networks (RNN) and LSTM can be viewed as instances of the state space framework. In our formulation, we model metrics using structural time series models, which are again instances of state space models. Structural time series models are a family of models where the time series model is a combination of subcomponents. Each component represents a characteristic of the time series: a trend component, a seasonality component, an error term etc. The observable $y_{t}$ is then constructed as 

\begin{equation}
y_{t}=u_{t}+\gamma_{t}+\epsilon_{t},
\end{equation}

where $u_{t}$, $\gamma_{t}$, and $\epsilon_{t}$ are the trend, seasonal, and error terms respectively. Note, formulating the models as structural time series models allows us to systematically model each feature of the time series and build complex models. The specific choices we consider for the three terms will be discussed in the next section.  Having chosen a form for the three components, the model is cast in the state space form. Recasting all models in the state space form allows us to use a common approach for the fitting of parameters and prediction of future observations. In particular, linear Gaussian state space model makes use of the Kalman filter for parameter estimation and prediction. We use the statsmodels package for model implementations \cite{statsmodels}.

One observation we made when working with time series metrics in the domain is that many metrics related to revenue and user engagement often vary exponentially with time. When compounded with the seasonality of these metrics, we find that the variation in their values increases with time. To deal with such cases, we use data transformation, a commonly used technique in the time series domain to transform nonlinear data to make them amenable to linear modeling. In particular, we test logarithmic transformation of the time series. Once transformed, we apply the same structural time series model as discussed above.

\subsection{Model selection}
To ensure generalizability of our framework across different types of time series, for each metric that is configured the framework evaluates a range of models on their ability to fit the data. We consider a suite of classical time series models, each constructed as a combination of a trend, a seasonal, and an error component. The different trend, seasonal, and error components we consider for the models are listed in Table~\ref{tab:classical}

\begin{table*}
\centering
  \caption{List of classical model components considered}
  \label{tab:classical}
  \begin{tabular}{cc}
    \toprule
    Component & Types considered \\
    \midrule
    Trend & linear model, local level, local linear\\
    Seasonal & hourly, daily \\
    Error & Gaussian, AR(p): autoregressive model of order p=1,2 \\ 
  \bottomrule
\end{tabular}
\end{table*}

Each metric that is configured on our system goes through a model selection step where all models obtainable by considering various combinations of trend, seasonality, error from Table~\ref{tab:classical}  are evaluated for their prediction capability. As mentioned previously, we consider an 80:20 split of the historical data of the metric to train and evaluate each model. The model with the lowest prediction error is then promoted for subsequent anomaly detection. The models we consider are flexible to adapt to changing trend and seasonality, and do not necessitate frequent training. However, in a production setting, it is important to keep tracking the performance of the selected model and to re-run the model selection procedure when the performance of the model starts to deteriorate significantly.

\subsection{Anomaly detection}
The anomaly detection module detects anomalies as new data points come into the system. The metric-specific model that was developed through model selection is used to predict the expected value of the metric at each time stamp. In the classical time series models, due to the linear Gaussian form of the models, the predicted uncertainty of the observations is also available through closed form updates. The predicted observation value takes a Gaussian distribution. We use this Gaussian distribution and the observed value of the metric to determine whether an observation is an anomaly or not. As is commonly done in the field, we use a k-sigma rule to determine if an observation is an anomaly or not. If a metric observation deviates more than k-sigma from its expected value, the observation is deemed to be anomaly.

\begin{figure*}
\begin{subfigure}{\textwidth}
 \centering
 \includegraphics[width=.8\linewidth]{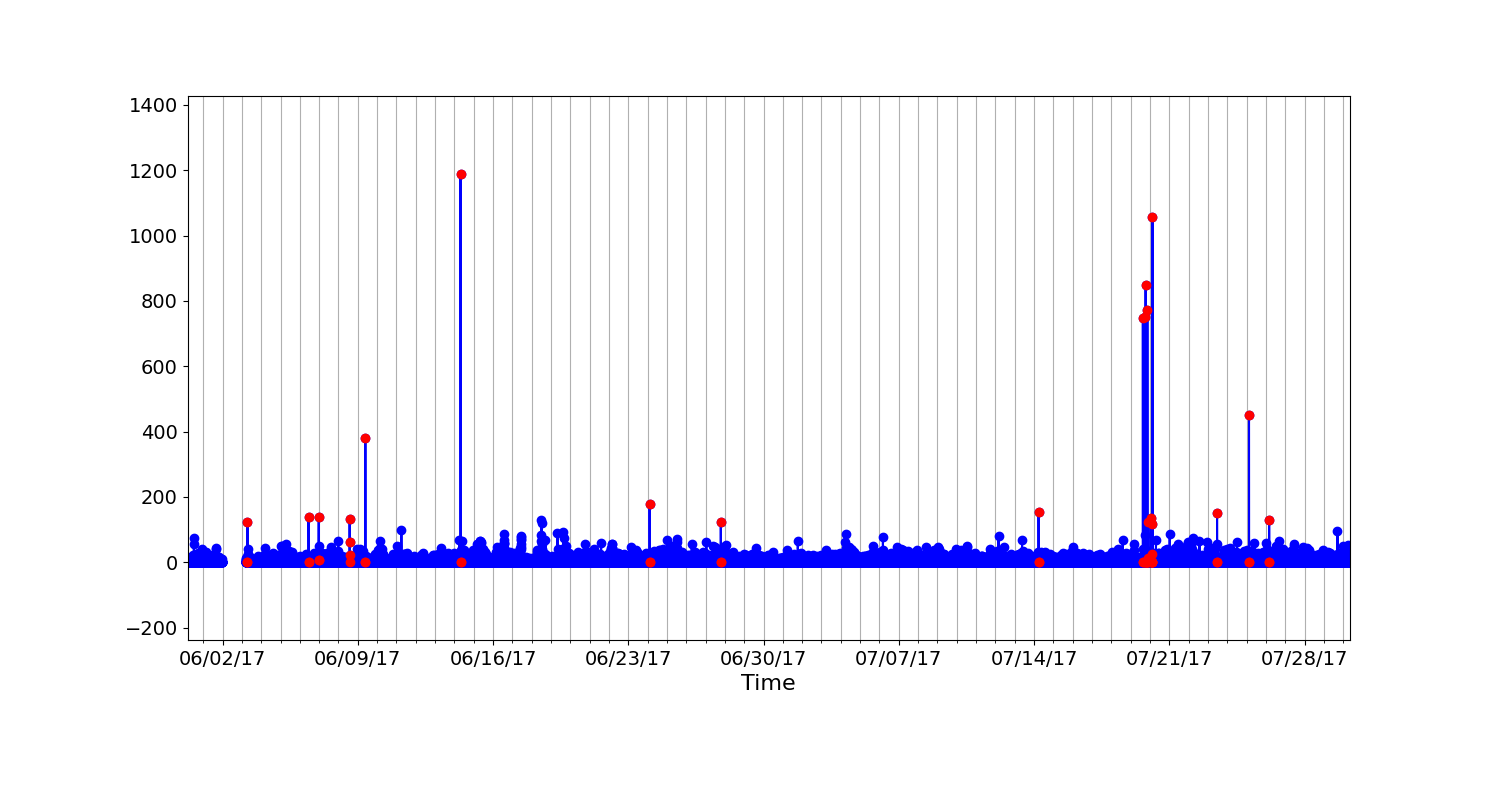}
  \caption{Time series 1}
  \label{fig:37}
\end{subfigure}%
\\
\begin{subfigure}{\textwidth}
\centering
\includegraphics[width=.8\linewidth]{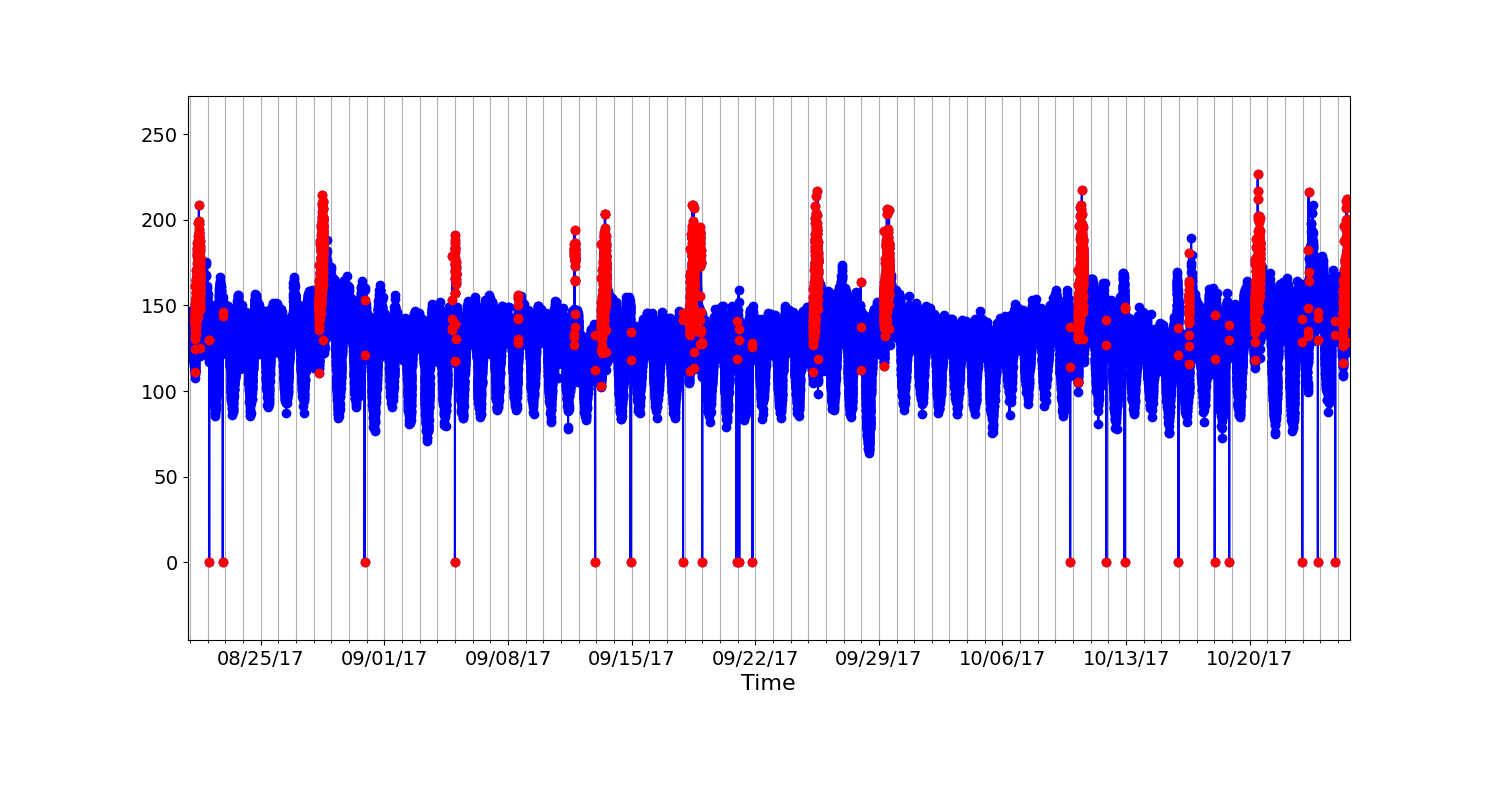}
\caption{Time series 2}
\label{fig:24}
\end{subfigure}%
\caption{Time series with mixed continuous and discrete values}
\end{figure*}

One observation we made when performing anomaly detection on metrics in the internet-services domain was that in many cases the metrics would take a mixed continuous and discrete values. For example, the error rate of requests to a server is usually zero but on occasion spikes to a non zero value. In this case, both a zero and nonzero value are considered expected unless the error rate becomes significantly high (Figure \ref{fig:37}). Similarly, request count to a server is normally nonzero but occasionally drops to zero. Here, a zero request count is usually a sign of system failure (Figure \ref{fig:24}). Using only a continuous value time series model like the linear Gaussian state space model typically leads to poor results in these settings. To better deal with the mixed scenario, we consider the discrete and continuous values separately. For the discrete value---in the above examples, a value of zero or not---we use a simple proportion model. We determine the proportion of zeros in the training data. If the proportion of zero is more than 1$\%$, we consider the zero value data points to be normal. If the proportion of zero data points is less than 1$\%$, we predict all zero value data points to be anomalies. For the continuous (non-zero) data points, we continue to use the approach described in the previous section. 

\subsection{Interpretability}
Besides accurate forecasting and modularity, one of the key benefits of structural time series models is that they are very easily interpretable. Because the time series are modeled as a combination of subcomponents representing features such as the time series trend, seasonality, and residual error, alerts can be quickly examined and decision be made whether to further investigate it. When the model generates an alert, in addition to the metric details, additional information such as the severity of the anomaly and full details regarding the model structure can be sent to the recipients. The model being modular, allows the receiver of the alert to carefully examine the nature of trend, seasonality, and error models used to arrive at the anomaly decision, and determine whether the alert needs further investigation. Overall this results in large savings in time invested by engineers and analysts to debug and root cause anomalies on a daily basis. 

\section{Experiments}
We run experiments to evaluate our framework for accuracy of anomaly detection against other commonly used approaches. 

\subsection{Datasets}
We use a publicly available dataset KPI from the internet-services domain to evaluate our framework. KPI is often used to evaluate time series anomaly detection approaches. The datasets consist of a range of times series of different characteristics and was released as part of the AIOPS data competition \cite{aiopsdataset}. It consists of 58 time series with anomaly labels collected from various internet-services companies, including Tencent, eBay etc. The time series have minutely or 5-minutely granularity.

\subsection{Metrics}
In our experiments, we evaluate our framework in terms of the precision, recall, and F1-score of anomaly detection. Typically, users of anomaly detection systems are not interested in point-wise detection of anomalies. Rather, one is more interested in detecting each incident of anomalous event, since an anomalous event generally manifests in a contiguous series of data points being labeled as anomalies. As a result, it generally suffices if the anomalous event is identified. However, it is important that the anomalous event is detected without a significant delay. These criteria have led to the use of a modified version of precision, recall, and F1-score measures to evaluate anomaly detection systems in the internet-services domain \cite{alibaba2018,microsoft2019}. A contiguous anomaly series is considered to be correctly identified if the model detects an anomaly within a delay of k points rom the start of the series. If an anomaly is not identified within a delay of k points, the entire contiguous anomaly series is considered a false negative.

\begin{figure*}
\centering
\includegraphics[width=0.7\textwidth]{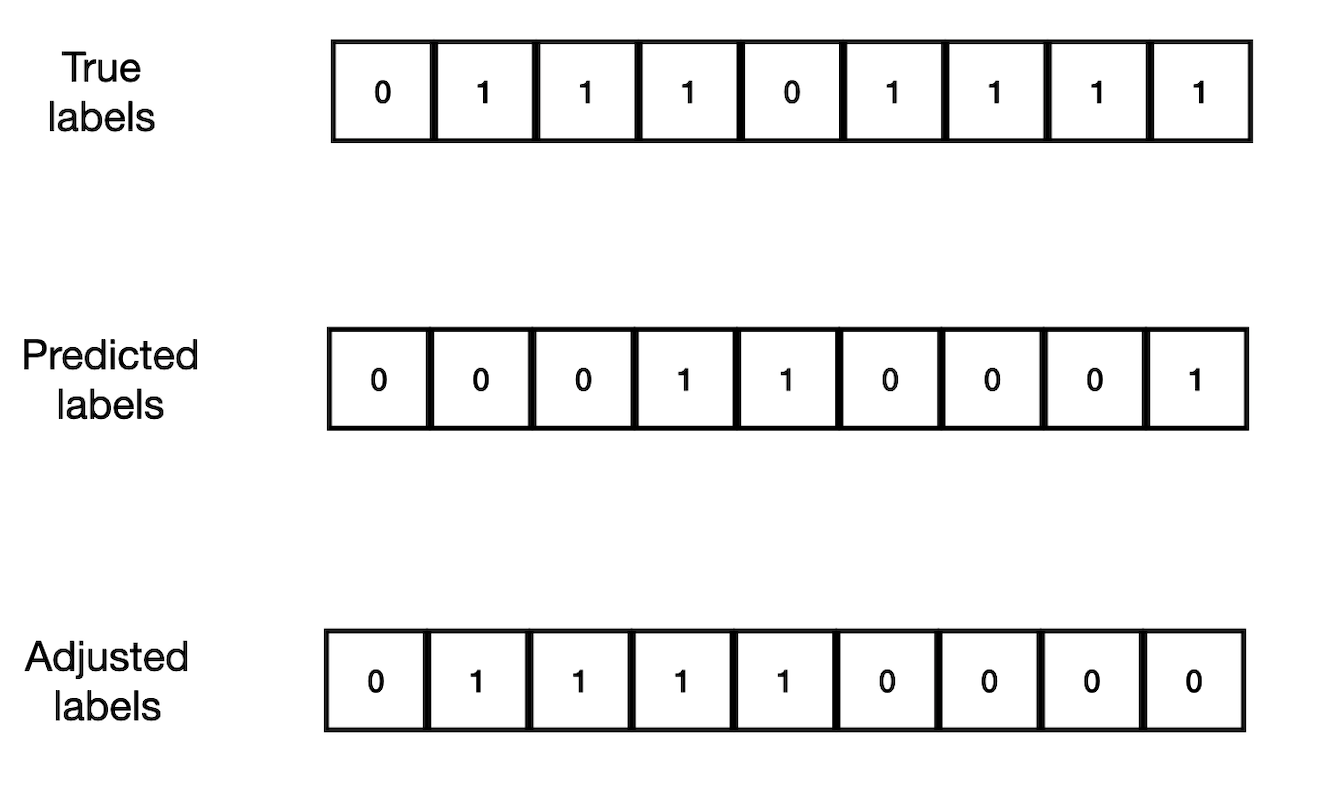}
\caption{Illustration of the evaluation strategy for a permitted delay of 2 data points; the first series of anomalies is identified within a delay of two points hence entire series is considered to be detected. The second series is not detected within a delay of two and as a result all points in the series are considered predicted normal}
\label{fig:evaluation_strategy}  
\end{figure*}

The evaluation strategy is shown in Figure \ref{fig:evaluation_strategy}. We see that there are two anomaly series. Let's say the permitted delay is 2 points. The predicted anomalies are shown in the second row. The first anomaly series is correctly identified within a delay of two points and thus the entire series is considered to be detected as anomalies. The second anomaly series is detected with a delay greater than 2 data points and hence the entire series is considered to be normal. The adjusted anomaly predictions are shown in the third row and are used for the estimation of precision, recall, and F1-score.

\subsection{Setup}
For a fair comparison our framework to other commonly used anomaly detection approaches, we setup the evaluation following \cite{microsoft2019}. We use the time series data points from the first half of each time series to train the model and test the model on the second half of the series. Note, since our approach is unsupervised, no anomaly labels are used for training or predictions. The anomaly labels in test set (second half of the time series) are used to compute precision, recall, and F1-score values. In the case of the AIOPS dataset, since the time series can be very large, we use a maximum of two weeks of data points prior to the test set for training the model. 

\begin{table}
  \caption{Results of AIOPS dataset}
  \label{tab:aiops}
  \begin{tabular}{lccc}
    \toprule
    Threshold&Precision&Recall & F1-score \\
    \midrule
    3-sigma & 0.75 & 0.83 & 0.79\\
    4-sigma & 0.89 & 0.79 & 0.84\\
    5-sigma & 0.94& 0.73 & 0.82 \\ 
    6-sigma & 0.96& 0.68 & 0.80 \\ 
  \bottomrule
\end{tabular}
\end{table}

\begin{table*}
\caption{Comparison of anomaly detection results}
\label{tab:comparison_aiops}
\centering
\begin{tabular}{|l|ccc|}
\toprule
& \multicolumn{3}{|c|}{AIOPS} \\
\midrule
Model &Precision&Recall & F1-score \\
\midrule
FFT & 0.48 & 0.61 & 0.53 \\
Twitter - AD & 0.41 & 0.28 & 0.33 \\
Luminol & 0.31 & 0.65 & 0.42 \\    
DSPOT & 0.62 & 0.45 & 0.52 \\
DONUT & 0.37 & 0.33 & 0.35 \\
Microsoft SR+CNN & 0.80 & 0.75 & 0.77 \\
Our framework & 0.89 & 0.79 & {\bf 0.84} \\
\bottomrule
\end{tabular}
\end{table*}

\subsection{Results}
The results of our model on the AIOPS dataset at different threshold settings is shown in Table~\ref{tab:aiops}. As expected, as the threshold for anomalies increases from 3-sigma to 6-sigma, the precision of anomaly detection increases while the recall goes down. Table~\ref{tab:comparison_aiops} shows the application of other unsupervised anomaly detection approaches to the AIOPS dataset \cite{microsoft2019}. We find that the F1-score of our framework is significantly better than the best reported F1-score on the AIOPS dataset. The computational efficiency of our approach is also very high since the evaluation of each data point is through sequential closed form updates resulting from the linear Gaussian structure of the models. In addition to the high accuracy of our framework, it allows for easy interpretation of the model structure, enabling rapid decision-making with regard to whether an anomaly needs to be investigated, thus saving an enterprise many valuable man-hours that are frequently spent investigating false alarms. 

One important consideration to keep in mind while using time series anomaly detection in production setting is that continuous monitoring of the model quality is also critical. Businesses release new features in order to satisfy customer demand and grow market share. With frequent deployment of new features, sometimes the characteristic of the time series metrics can change with time. This can make the model trained previously inapplicable and necessitates fresh training of the model. Continuous assessment of performance is thus important to prevent deterioration of model quality and ensure anomaly detection results are reliable.

\section{Conclusion and Future Work}
This paper presents a novel framework for time series anomaly detection in the internet-services domain. It shows that through a combination of structural time series modeling, data transformation, model selection, and mixed continuous-discrete anomaly detection, one can obtain high performance anomaly detection systems. The core components of our anomaly detection framework are a time series modeling component, a model selection engine, and an anomaly detection module. Experiments on the AIOPS datasets show the accuracy of our framework is higher than existing unsupervised time series anomaly detection approaches. In addition, our framework provides easy interpretation of the underlying time series model, thus enabling rapid decision-making as to whether an anomaly deserves further investigation.

Future work could involve exploring neural network models that would help capture more complex nonlinear dependencies while continuing to be interpretable. In addition, it would be interesting to explore ways to use the framework proposed in this paper to make the subsequent process of root-cause-analysis faster and more reliable.

\section{Code}
The code developed as part of this work can be found in the following repository: https://github.com/nikhilgalagali/adservice

\bibliographystyle{plain}
\bibliography{kdd2022}

\end{document}